\documentclass[10pt,a4paper, oneside]{article}
\usepackage{epsfig,subfigure,multicol,float}

\title{IR-based Communication and Perception in Microrobotic Swarms}
\author{S.~Kornienko, S.~Kornienko \\
{\small Institute of Parallel and Distributed Systems, University of Stuttgart,}\\
{\small Universit{\"a}tsstr.~38, D-70569 Stuttgart, Germany}
\thanks{Original paper: IROS 2005, WS on Task-oriented Mobile Actuator and Sensor Networks, Edmonton, Canada. Extended version appeared on the 7th Workshop on Collective \& Swarm Robotics, 18 November, University of Stuttgart, Germany, 2010}
}

\begin{document}

\date{}
\maketitle

\begin{abstract}
In this work we consider development of IR-based communication and perception mechanisms for real microrobotic systems. It is demonstrated that a specific combination of hardware and software elements provides capabilities for navigation, objects recognition, directional and unidirectional communication. We discuss open issues and their resolution based on the experiments in the swarm of microrobots "Jasmine".
\end{abstract}

\section{Introduction}

The important issue in swarm research \cite{Sahin04} is how a large number of collective agents is well coordinated, how do they work cooperatively on different tasks~\cite{Bonabeau99} ? We encounter here many issues, such as coordination rules~\cite{Luna00}, collective decision making~\cite{Kornienko_OS01} with analytic~\cite{Levi99} and algorithmic~\cite{Kernbach08} approaches, cooperative~\cite{Kornienko_S03A}, \cite{Kornienko_S04} and other planing approaches. More generally, these issues are related to different fields of AI~\cite{Kornienko_S05b} and designing of emergence~\cite{Kornienko_S04a}, \cite{Kornienko_S04b} e.g. through embodiment~\cite{Kornienko_S05e}. However all these mechanisms work only when robots are interacting. There are several ways of how the robots can interact: robots observe environment and the behavior of other robots, physical interaction, indirect interaction trough environment or they communicate. Since the microrobots are restricted in sensing and computation, the perceptive way of interaction has a limited application for microrobotic swarms. There are works on e.g. recognition of robots by emitted IR-radiation \cite{Kornienko_S06}, color perception \cite{Zetterstrom06} or even using collisions as interactions among robots, however complex cognitive~\cite{Kornienko_S05a} and behavioral mechanisms~\cite{Weiss99} are impossible without communication. Moreover, the developed in MAS community mechanisms of rule-based coordination~\cite{Durfee88}, e.g. token exchange~\cite{Scerri05}, assume that robots can exchange semantic information. 

Developing communication mechanisms for real microrobotic swarm~\cite{Kornienko_S05d}, we encounter a few problems of technological and methodological character. First of all, robots have only limited communication radius. This allows avoiding the problem of communication overflow in large-scale swarms (100+ robots), however opens the problem of propagating the relevant information over the swarm. This information concerns e.g. energy resources, behavioral goals, dangers and so on. Robots are restricted in hardware for using algorithms and protocols known in the domain of distributed systems~\cite{Coulouris01}. Therefore new concepts and new protocols should be developed for the swarm communication. Not only software protocols, but also communication hardware should be adapted to the need of large-scale swarms. This primarily concerns a multi-channel equipment for omnidirectional local communication, using of low-level signals, optimization of the emitted energy and solution of routing problem. We say that only a good interplay between hardware, software and robots behavior allows a reliable information transfer in a swarm.

The goal of this paper on the one hard is to overview the original design of the IR-based mechanisms of perception and communication for the Jasmine robot, represented in 2004-2006 in such confreres and accompanying workshops as IROS, AMS, IJCAI and ECAI. On the other hand we need to revise these ideas for the later development from 2007-2010 between swarm and artificial organisms~\cite{Kornienko_S07}, evolutionary robotics~\cite{Kernbach09Platform}, online embodied evolving~\cite{Kernbach08online}, \cite{Schlachter08}, systems capable of structural modifications~\cite{Kernbach08_2} and more general fields of adaptive systems \cite{kernbach09adaptive}, \cite{Kernbach10IntSys}. This should facilitate further elaboration of ideas for embodied cognition on the workshops on collective \& swarm robotics.

The rest of the paper is organized in the following way: Sec.~\ref{sec:ir-based} is devoted to hardware mechanisms of IR-based perception, and Sec.~\ref{sec:scaning} -- to object recognition based the described hardware. Sec.~\ref{sec:embodiment} deals with communication issues. Finally, Sec.~\ref{sec:conclusion} concludes this work.

\section{IR-based perception}
\label{sec:ir-based}

\begin{figure}[h]
\centering
\includegraphics[width=0.5\textwidth]{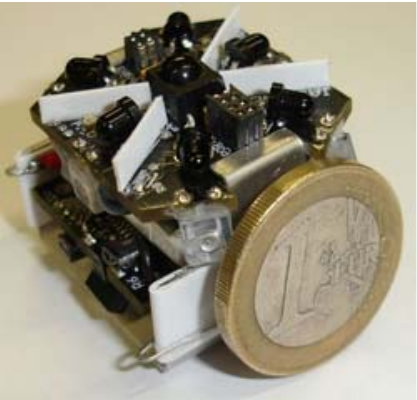}
\caption{\small The "Jasmine" microrobot \label{fig:robot}}
\end{figure}

The "Jasmine" microrobot, shown in Fig.~\ref{fig:robot}, 
is a public open-hardware development at
\textbf{www.swarmrobot.org}, measures $30\times30\times20 mm^3$ in size and has two small DC motors with an integrated planetary gearbox. The microrobot has two circuit boards, the motor board and the main board, which communicate via a 200 kHz I2C interface. The main board holds an ATmega 168 microcontroller, six (60$^\circ$ opening angle) IR channels (used for proximity sensing and communication) and one IR geometry-perception-channel (15$^\circ$ opening angle). The sensing area covers a 360$^\circ$ rose-like area with maximum and minimum ranges of 200mm and 100mm respectively~\cite{Kornienko_S05d}. The physical communication range can be decreased through a change of sub-modulation frequency. The main board also supports remote control, differential light sensing, energy management, ZigBee communication and is primarily used for the behavioral control of the robot and for upper extension boards. The motor board has an ATmega 88 microcontroller and is used for motor control, the odometrical system, energy control, touch (short-range reflective IR sensor), and color sensing; it also provides another four channels for further sensors/actuators.

\begin{figure}[h]
\centering
\subfigure[]{\includegraphics[width=0.45\textwidth]{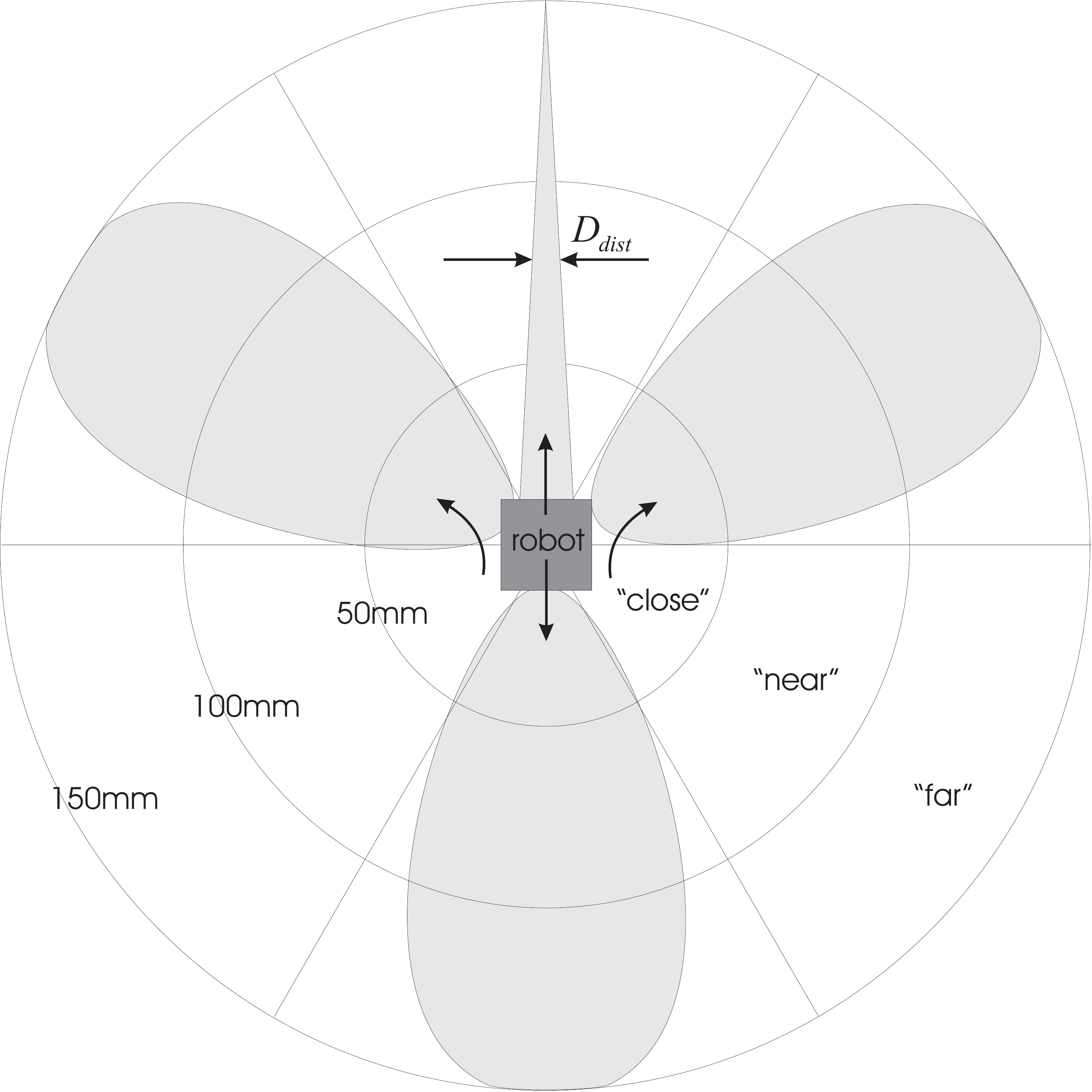}}~
\subfigure[]{\includegraphics[width=0.45\textwidth]{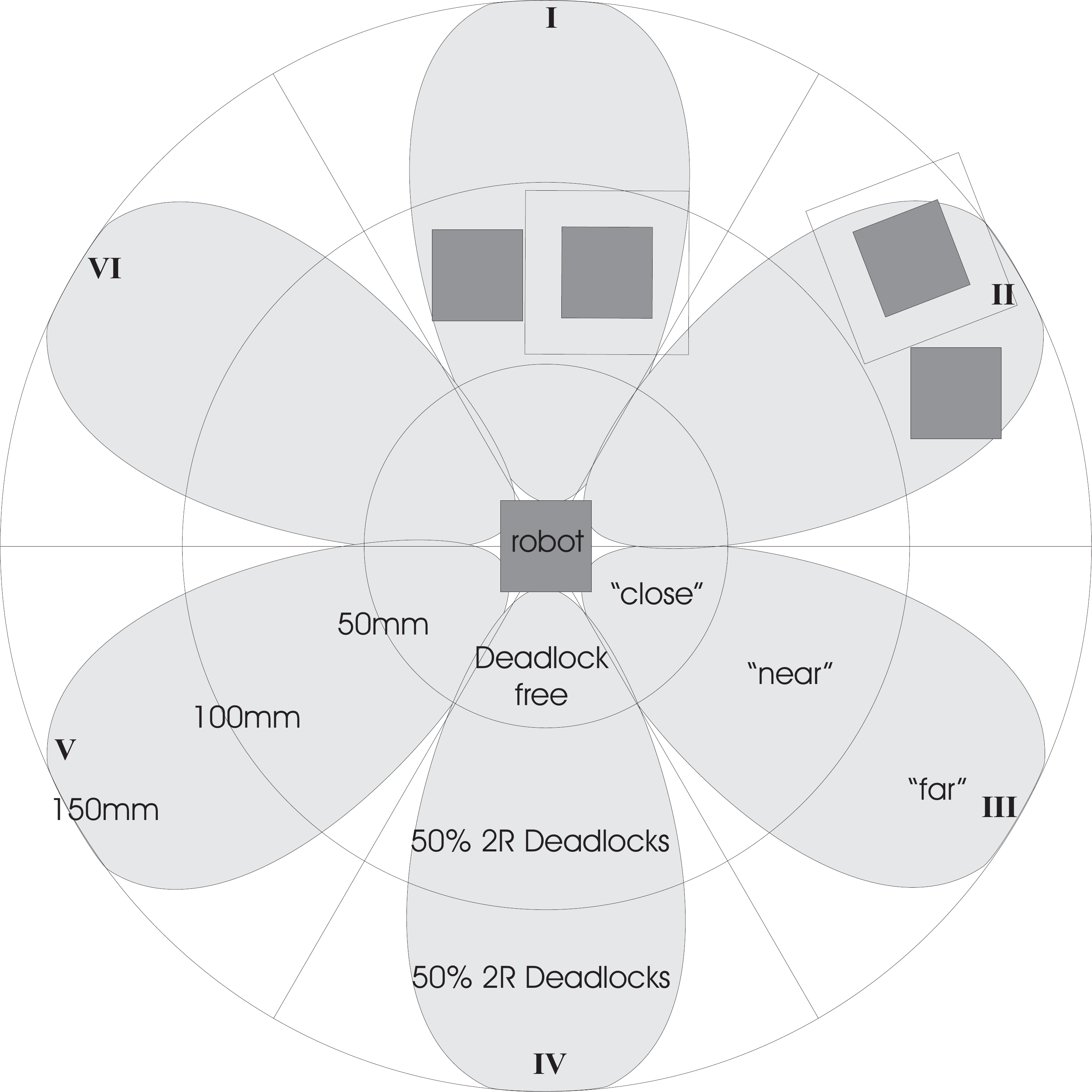}}
\caption{\small
{\bf (a)} Proximity and distance sensors in a micro-robot;
{\bf (b)} IR-sensors used for 6-directional communication.
\label{fig_sensors}
}
\end{figure}

Microrobot has IR-emitters and receivers for sensing its own environment. These IR-devices are used for a proximity sensing and obstacle detection, distance measurement and communication, see Fig.~\ref{fig_sensors}.
For the perception and objects recognition we use only the distance measuring sensor, so that we consider further only this sensor. For that we use a receiver with a wide opening angle and an emitter with as small as possible beam angle. In the microrobot we have how the Si phototransistor TEFT4300 (60$^o$, peak sensitivity 950 nm) and GaAs optical diode LD274 (radiant intensity 50 mW/sr, 20$^o$, 950 nm), all experiments are performed with this IR-pair. In further experiments we are going to use the high power GaAs/GaAlAs emitter TSAL5100 (radiant intensity $>$80 mW/sr, 20$^o$, 950 nm) and GaAs emitter with very narrow opening angle TSTS7100 (radiant intensity $>$10 mW/sr, 10$^o$, 950 nm).

\begin{figure}[h]
\centering 
\includegraphics[width=0.6\textwidth]{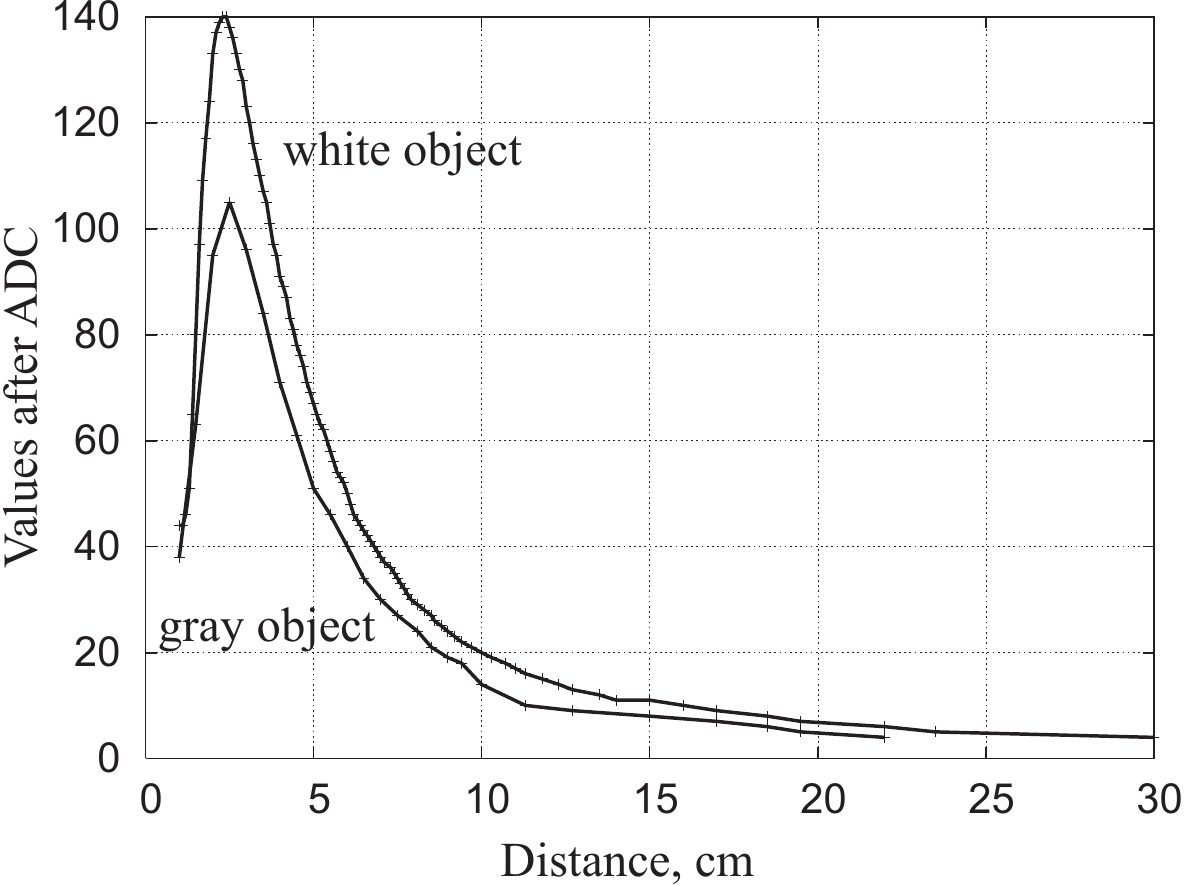}
\caption{\small Dependency between ADC values of emitter voltage on phototransistor and the distance to reflecting object. Shown are values for the white reflecting object (white paper) and the grey reflecting object (grey cardboard). \label{fig_distance} }
\end{figure}

The principle of object recognition is simple. As soon as a robot detects (by means of proximity sensors) an obstacle in the front of itself, it switches on the high power IR-emitter and after 1ms delay (needed to get reliable reflecting light) measures voltage on the emitter of phototransistor. The dependence between emitter voltage (after ADC) and the distance to an object is shown in Fig.~\ref{fig_distance}. Generally, this sensor perceives distances up to 300mm (in combination with TSAL5100 even more). However accuracy of measurement is different. In the area between 20mm and 70mm accuracy is of 1mm, from 70mm till 120mm accuracy is 3-5mm, 120mm - 200mm accuracy is 10 mm and after 200mm - 30-70mm. Therefore, the reasonable measuring distance for object recognition lies within 25mm-100mm (with the accuracy of 1-2mm). The reflecting light is also very sensitive to the color of reflection object. In Fig.~\ref{fig_distance} we demonstrate the distance measuring values for white and gray objects. Further in experiments we use only the objects of white color. The distance measuring also depends on the object's slope to a radiation ray. However, as marked in the performed experiments, this factor has minimal influence for the objects with convex or concave 90$^o$ corners. This kind of objects is also used further in experiments.

\begin{figure}[H]
\centering
\subfigure[]{\includegraphics[width=0.2\textwidth]{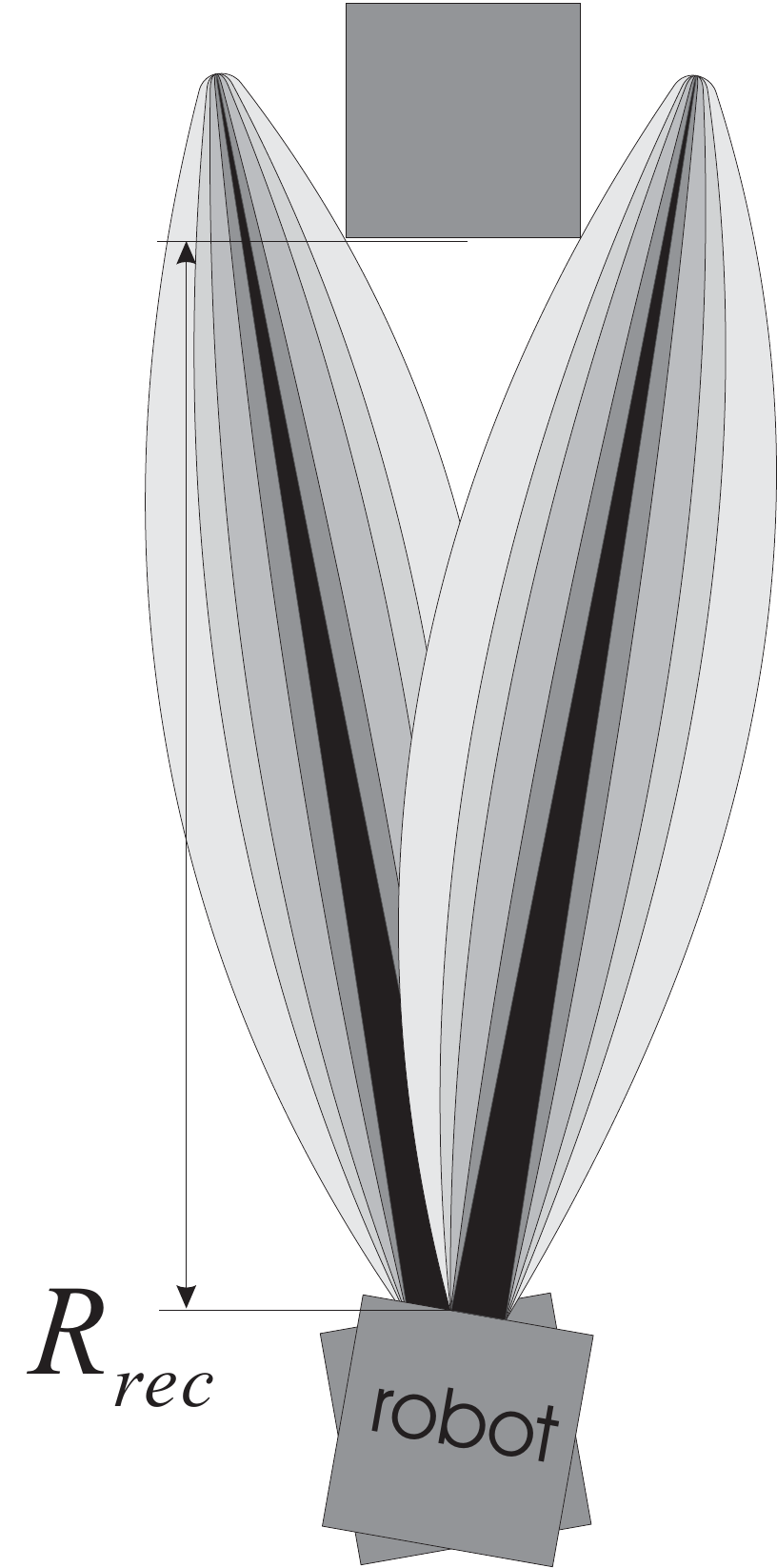}}~
\subfigure[]{\includegraphics[width=0.2\textwidth]{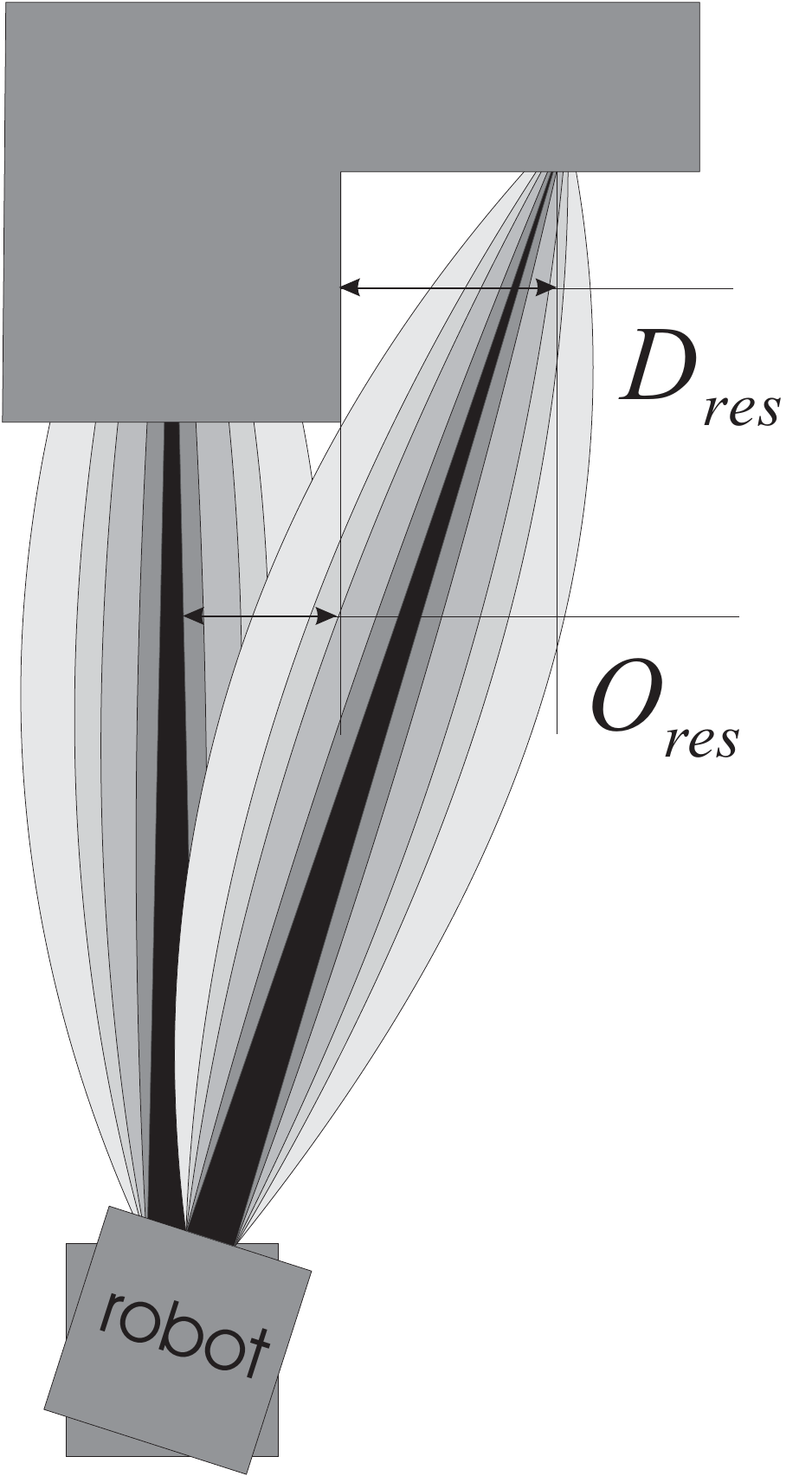}}
\caption{\small
{\bf (a)} Proximity and distance sensors in a micro-robot;
{\bf (b)} Problem of IR-interferences during communication.
\label{fig_scaning}
}
\end{figure}

Recognition of object is performed by scanning the object with IR-radiation ray, see Fig.~\ref{fig_scaning}.

\section{Experiments with Recognition Objects Geometries}
\label{sec:scaning}

\begin{figure}[h!]
\centering
\subfigure[\label{fig_object30mm}]{\includegraphics[width=0.47\textwidth]{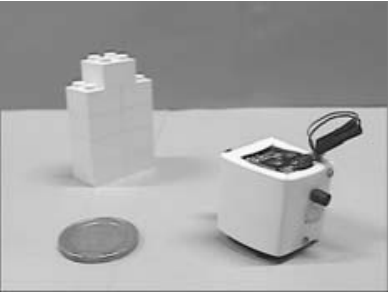}}~
\subfigure[\label{fig_object30mm1}]{\includegraphics[width=0.47\textwidth]{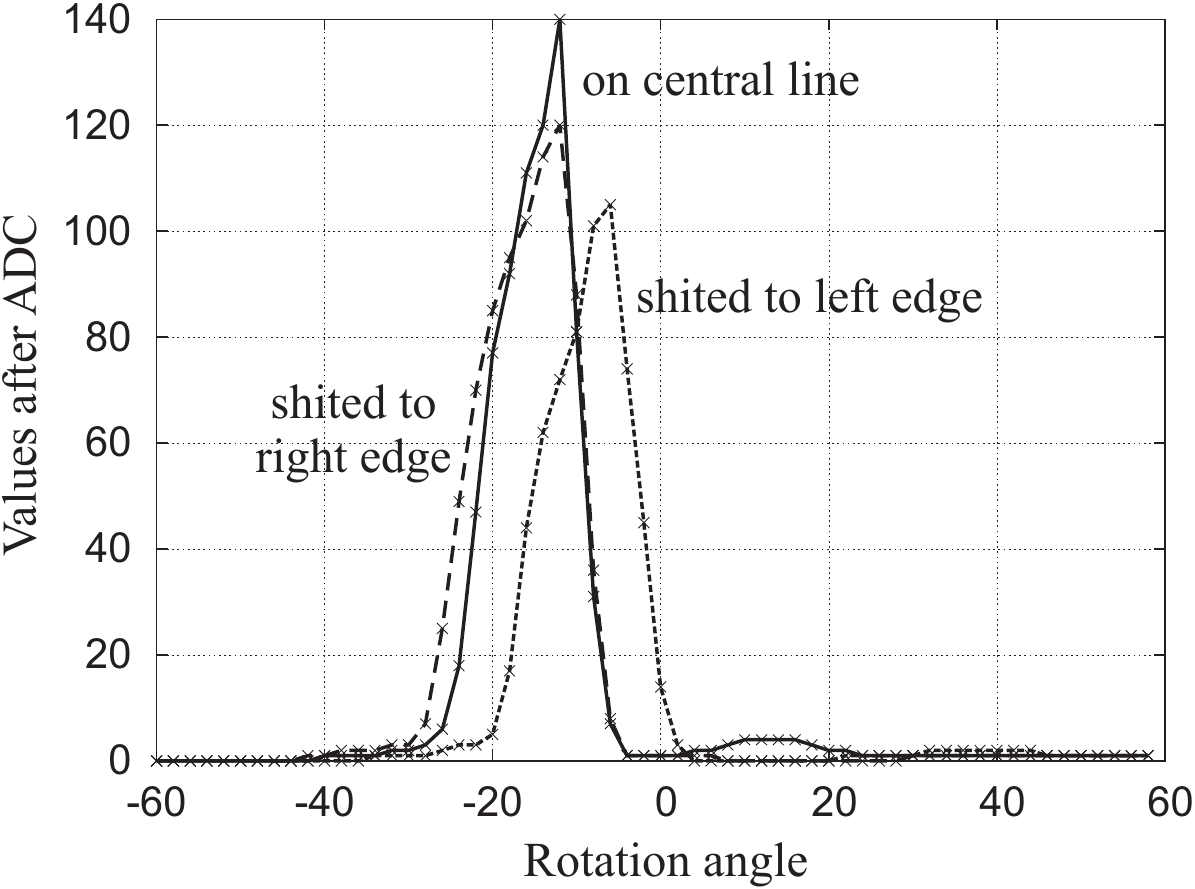}}\\
\subfigure[\label{fig_object50mm}]{\includegraphics[width=0.47\textwidth]{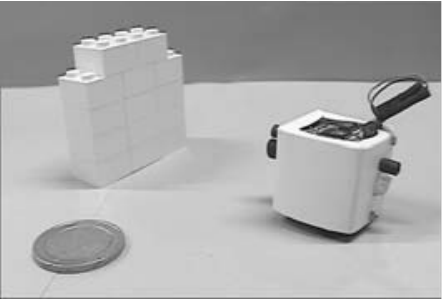}}~
\subfigure[\label{fig_object50mm1}]{\includegraphics[width=0.47\textwidth]{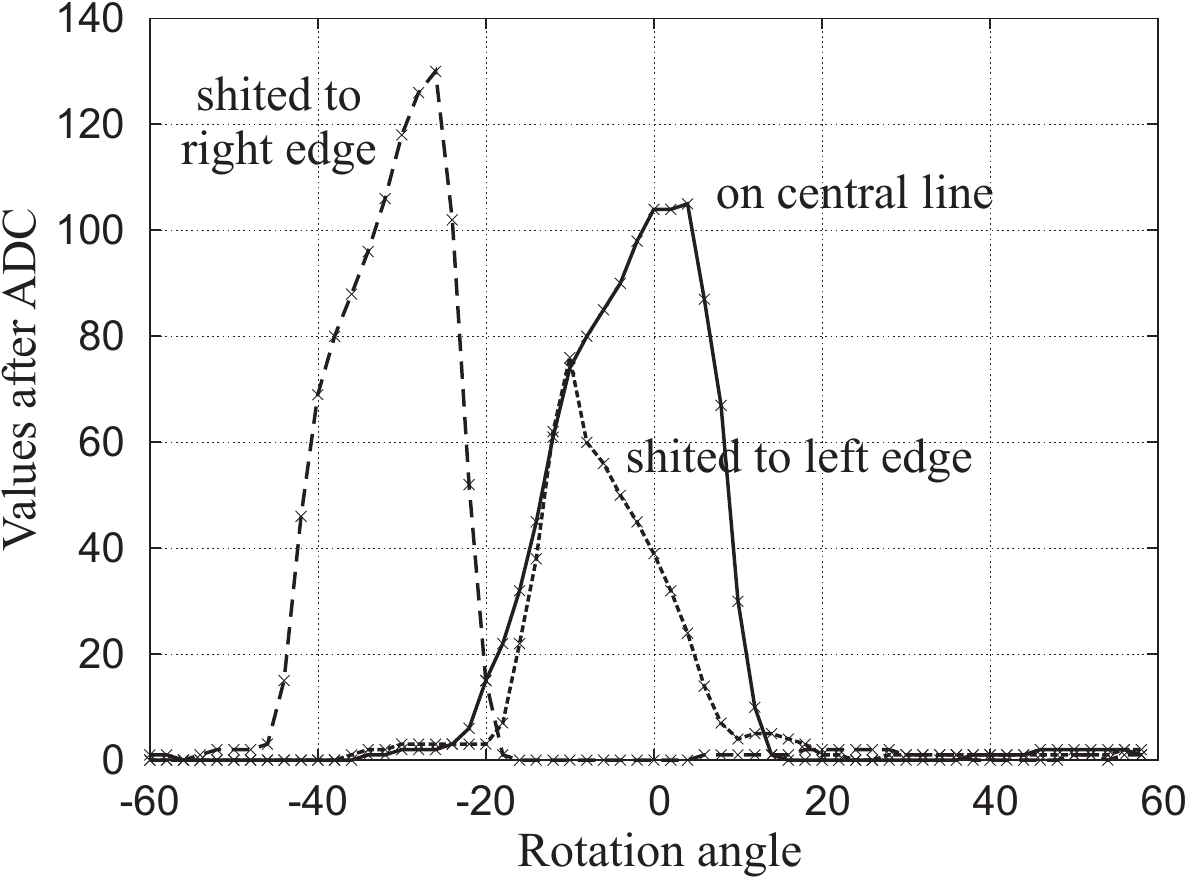}}\\
\subfigure[\label{fig_object144mm}]{\includegraphics[width=0.47\textwidth]{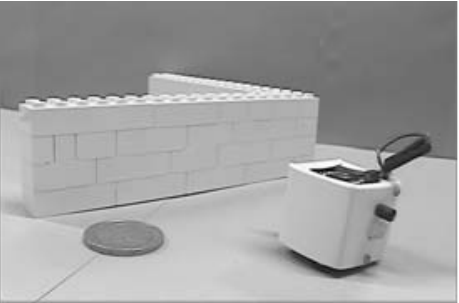}}~
\subfigure[\label{fig_object144mm1}]{\includegraphics[width=0.47\textwidth]{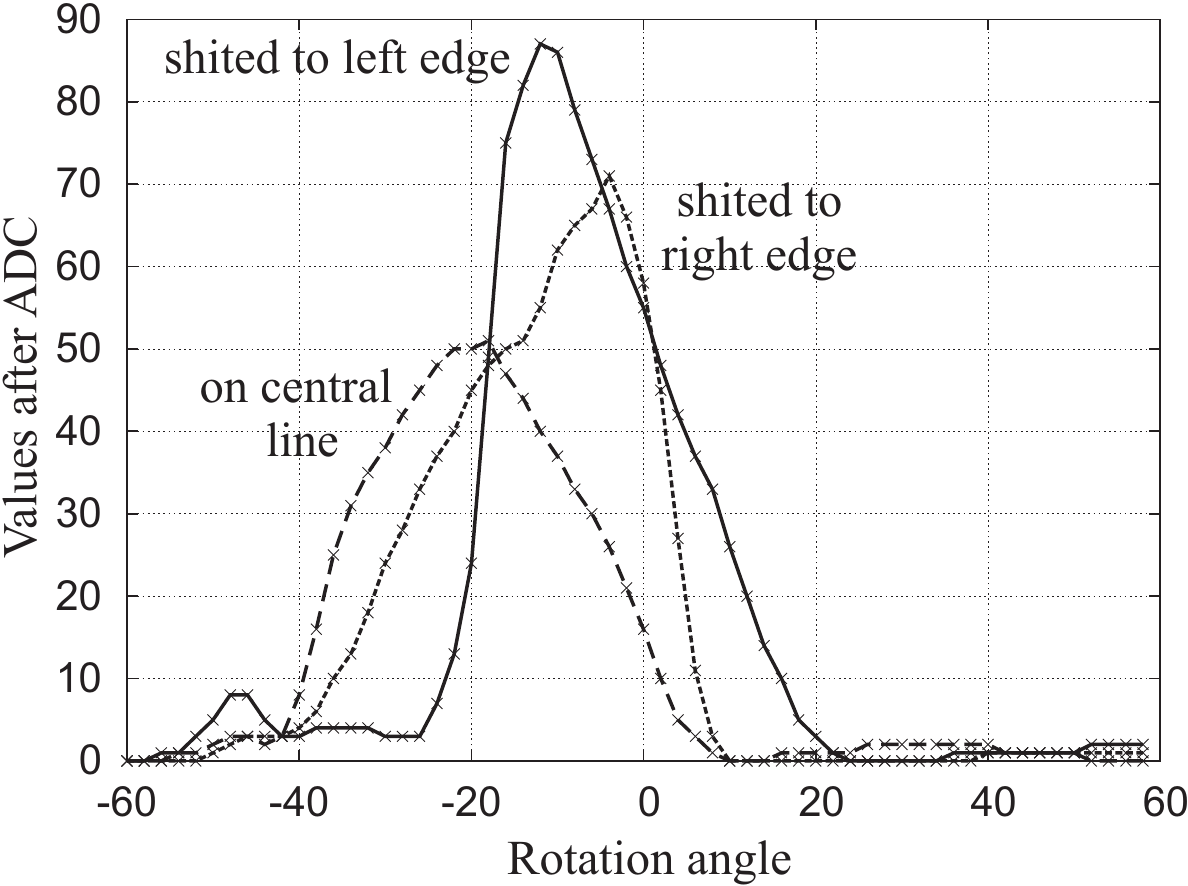}}
\caption{\small
{\bf (a)} Object 32mm;
{\bf (b)} ADC scanning values.
{\bf (c)} Object 48mm;
{\bf (d)} ADC scanning values.
{\bf (e)} Object 144mm;
{\bf (f)} ADC scanning values.
\label{fig:objectScan1}}
\end{figure}

\begin{figure}[htp]
\centering
\subfigure[\label{fig_corner_convex}]{\includegraphics[width=0.47\textwidth]{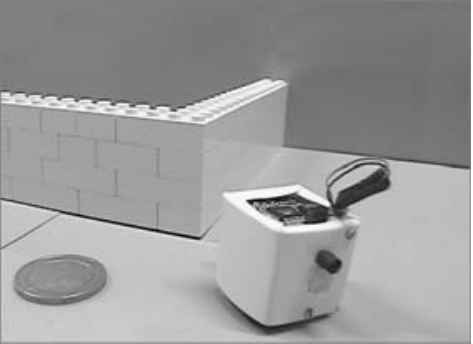}}~
\subfigure[\label{fig_corner_convex1}]{\includegraphics[width=0.47\textwidth]{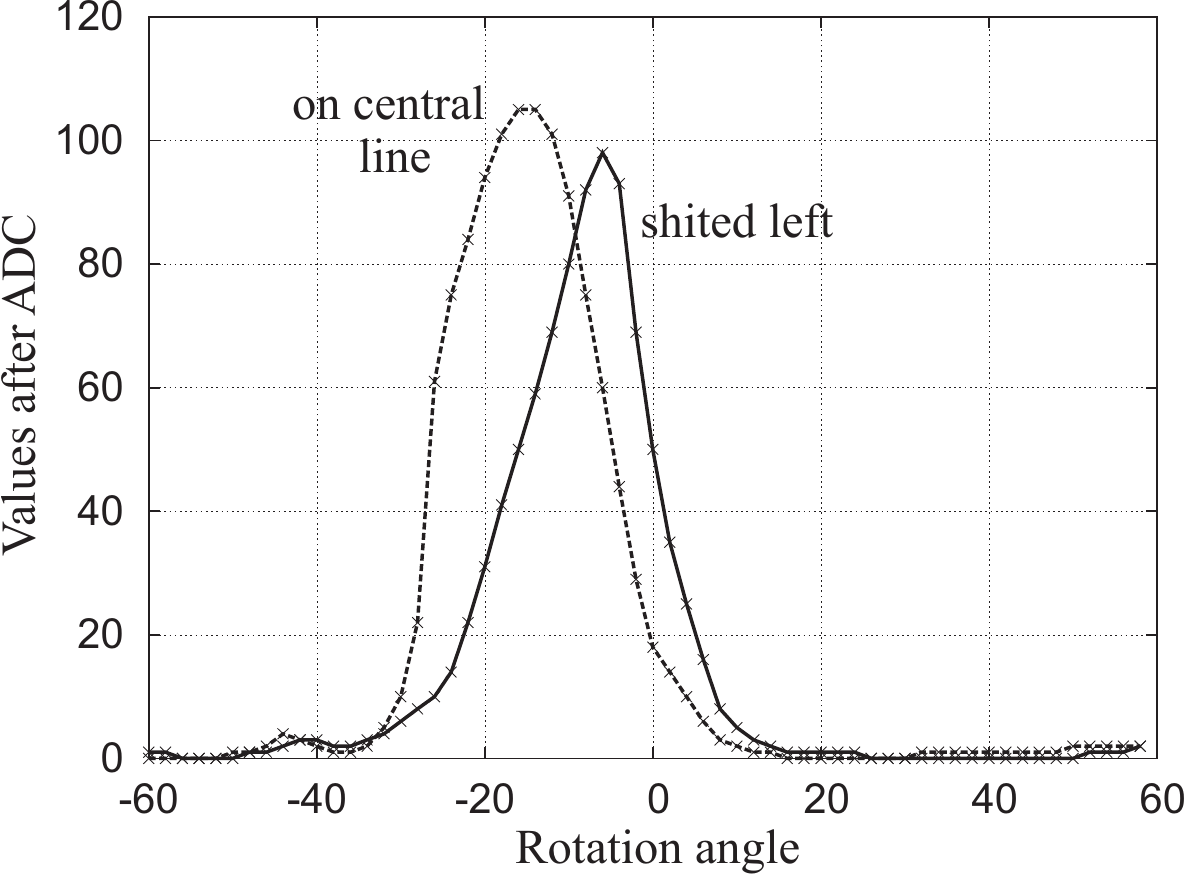}}\\
\subfigure[\label{fig_corner_concav}]{\includegraphics[width=0.47\textwidth]{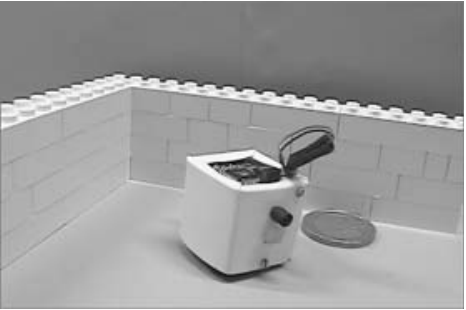}}~
\subfigure[\label{fig_corner_concav1}]{\includegraphics[width=0.47\textwidth]{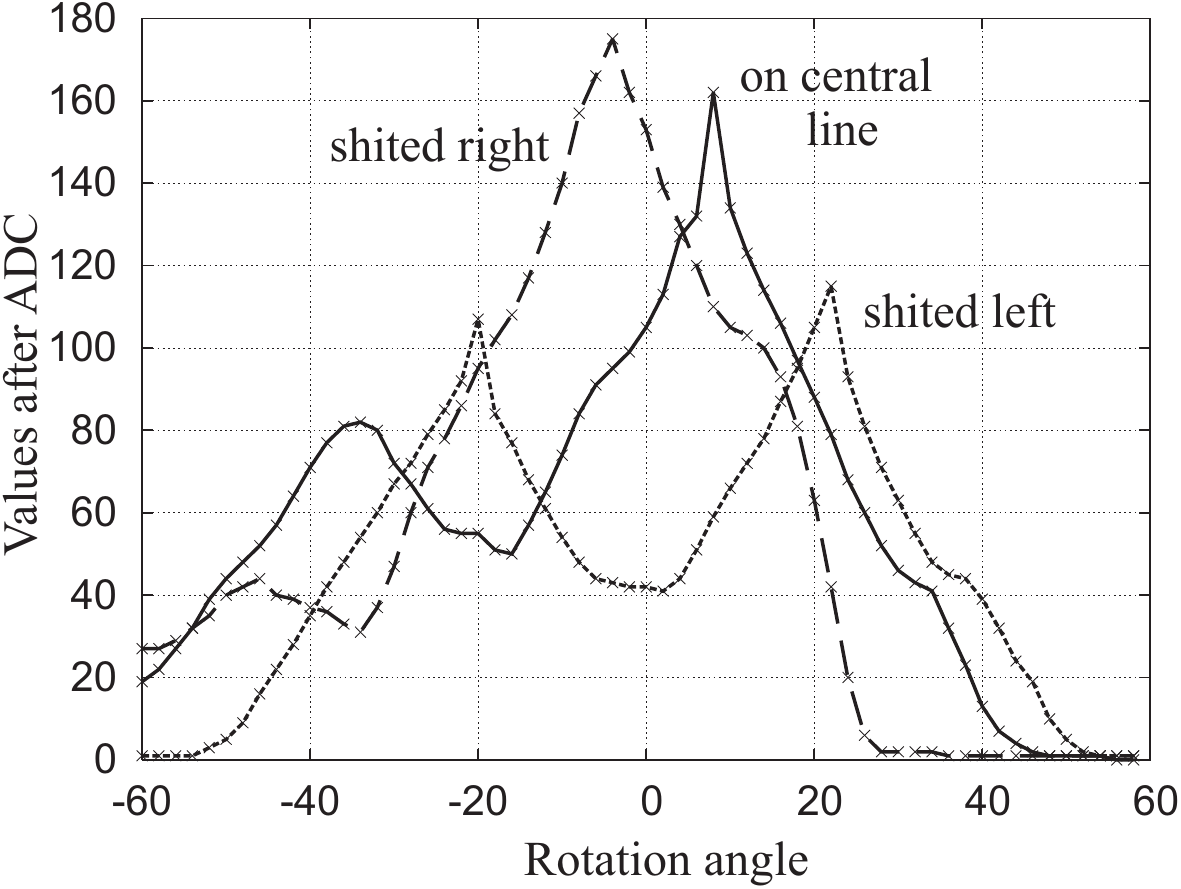}}\\
\subfigure[\label{fig_loch}]{\includegraphics[width=0.47\textwidth]{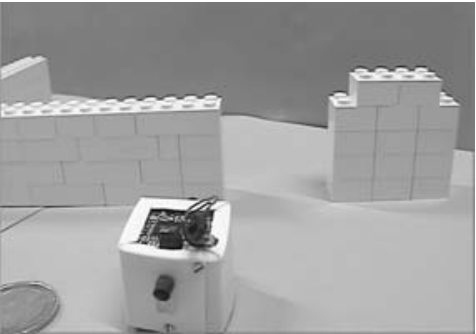}}~
\subfigure[\label{fig_loch1}]{\includegraphics[width=0.47\textwidth]{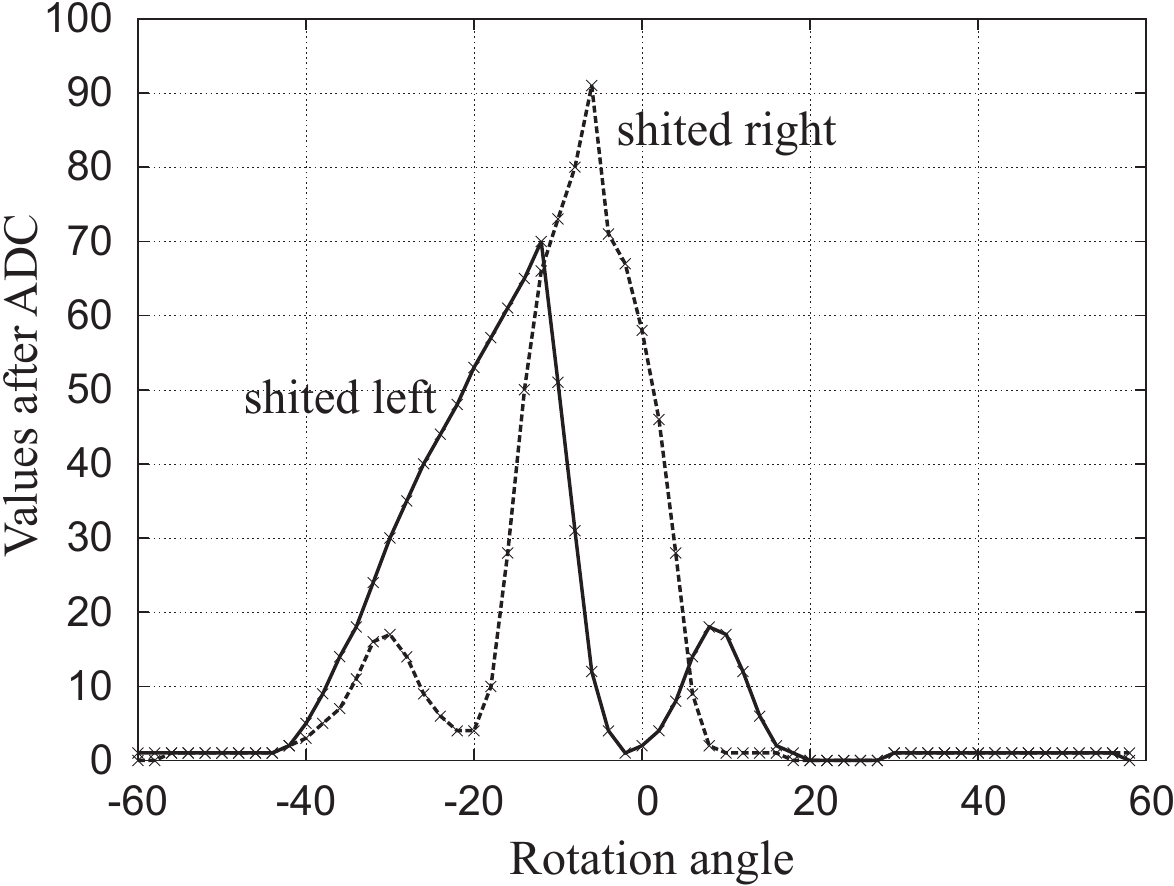}}\\
\subfigure[\label{fig_complex_geometry}]{\includegraphics[width=0.47\textwidth]{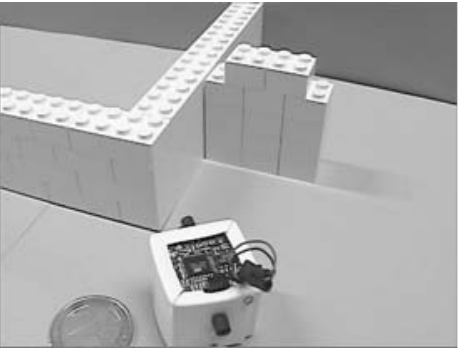}}~
\subfigure[\label{fig_complex_geometry1}]{\includegraphics[width=0.47\textwidth]{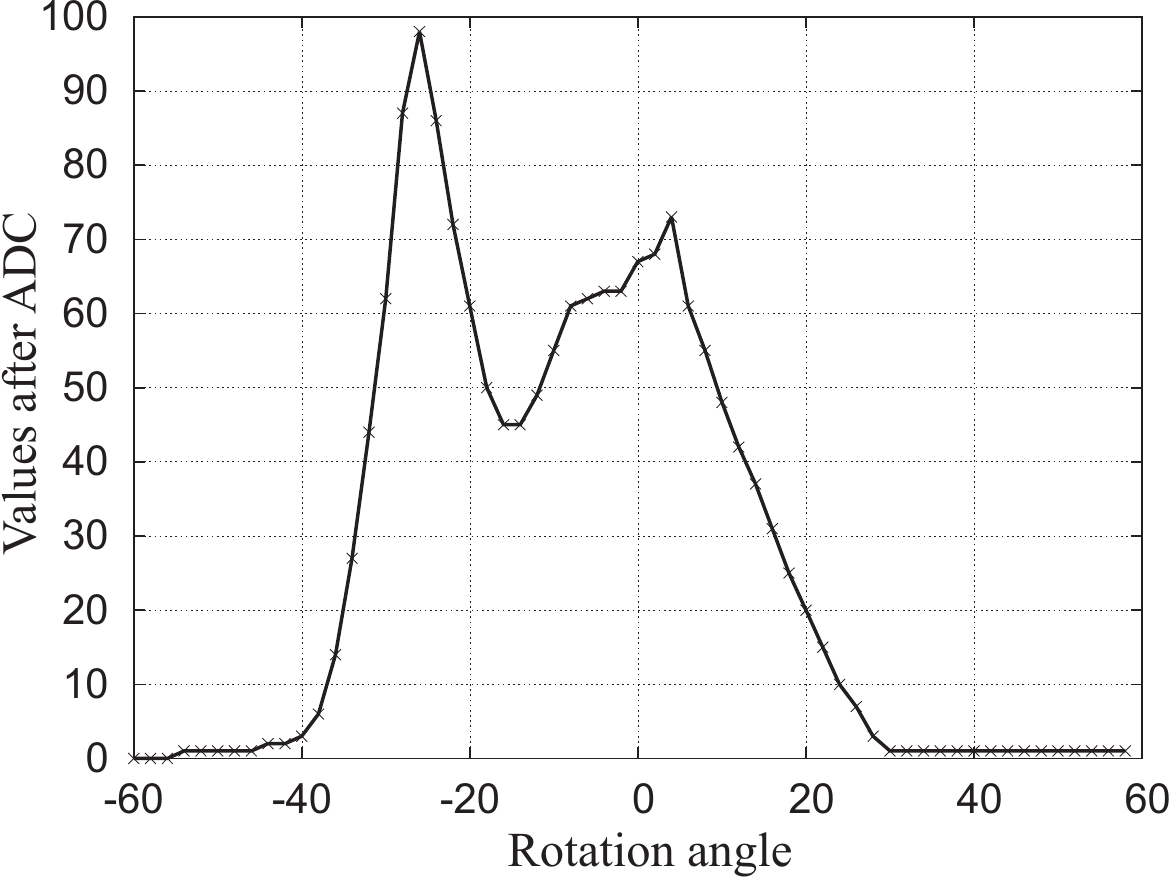}}
\caption{\small
{\bf (a)} Convex corner 90$^o$;
{\bf (b)} ADC scanning values.
{\bf (c)} Concave corner 90$^o$;
{\bf (d)} ADC scanning values.
{\bf (e)} 45mm hole between objects;
{\bf (f)} ADC scanning values.
{\bf (g)} Object with "complex geometry";
{\bf (h)} ADC scanning values.
\label{fig:objectScan2}}
\end{figure}

Experiments with the scanning objects geometries are shown in Fig.~\ref{fig:objectScan1}-  Fig.~\ref{fig:objectScan2}.

For perceiving the geometry of object, the resolution of the distance sensor is very important. The point is that in the center of radiation ray, the intensity if IR radiation is the highest. Closely to the bounds of this ray, the intensity of IR radiation becomes gradually degraded. The main component of a reflecting light consists of the energy of the cental radiation stream. However low-intensity "secondary streams" spread the reflecting light so that object's edges get non-recognizable from the surfaces, turned on some sharp angle. With a poor resolution of distance sensor, small geometrical elements cannot be perceived and so can not be used as features for recognition. The LD274 emitter in the microrobot has the opening angle of 20$^o$. The real opening angle is of 18-22$^o$. With this sensor the microrobot has the optical resolution of 25-30 mm for the distance of 100 mm (the minimal length of objects). Using the emitter TSTS7100 allows improving the optical resolution. However even with
LD274 a microrobot can perceives object of its own size on the border of recognition distance.

Not only the resolution of the IR-sensor is important for scanning the objects. During scanning, a microrobot turns on some degree. The more exact is the this turning, the more exact is the spatial resolution of sensor data. Microrobot does not possess any devices allowing to measure positions and orientation of body or wheels. Therefore there is only one way to rotate a robot, namely to turn the motors on and after some delay turn them off. This delay has to be so chosen, that a robot rotates on some fixed degree. The motors are controlled thought H-bridge SI9988, that can change a polarity of supplying current. Choosing normal polarity for one motor and inverse polarity for other motor, the robot can rotate just on one position (the sensor is placed on the chassis in 20mm away from the center of rotation, so that a small displacement of sensor still remain in the measurement).

In the experiments, when a robot detects an obstacle on the distance of 60mm $\pm$ 10mm, it stops and then rotates on the angle of 60 degree left. After that it scans the obstacle with the distance sensor by the rotation on 120 $^o$ right. During this scanning it writes the values of distances each 2 degrees into an integer array\footnote{In experiments the rotation left is 60$^o$,
however the rotation right is about 90$^o$, so that all diagrams in Figs.~\ref{fig_object30mm}-\ref{fig_complex_geometry} are shifted left. This different rotation angle is caused by the motion system of the microrobot, where wheels are not placed on one axis. However this specifics can be compensated by especial wheels and different driving voltage of each motor. This will be done in the next development of a chassis.}. In this way we have 60 values describing a visible geometry of the encountered obstacle. In
Figs.~\ref{fig_object30mm}-\ref{fig_complex_geometry} we demonstrate some geometries of obstacles and the scanned values of distances.

\section{Communication in a Swarm}
\label{sec:embodiment}

\subsection{Implementation}

\begin{figure}[htb]
\centering
\subfigure[]{\includegraphics[width=0.49\textwidth]{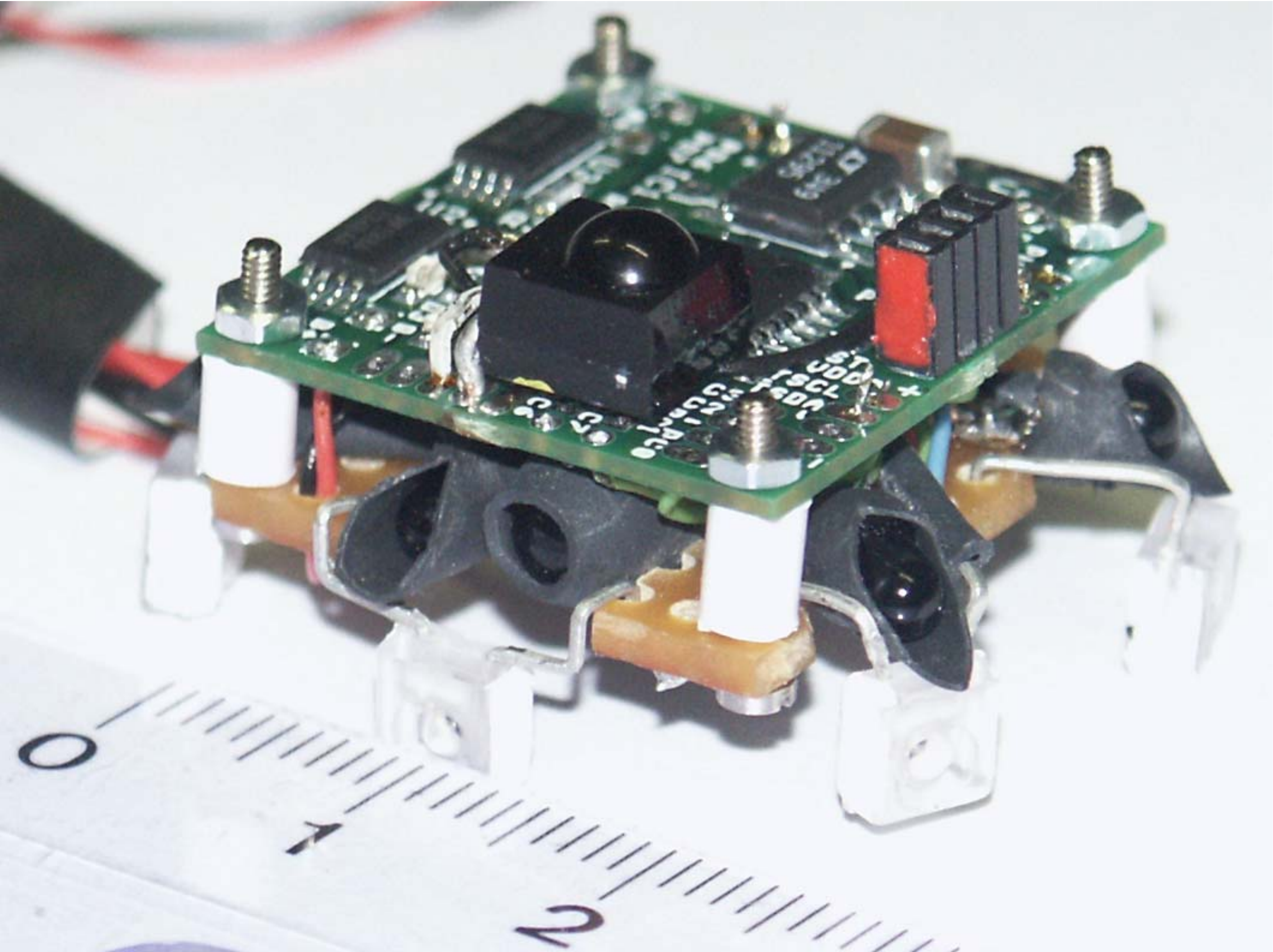}}~
\subfigure[]{\includegraphics[width=0.45\textwidth]{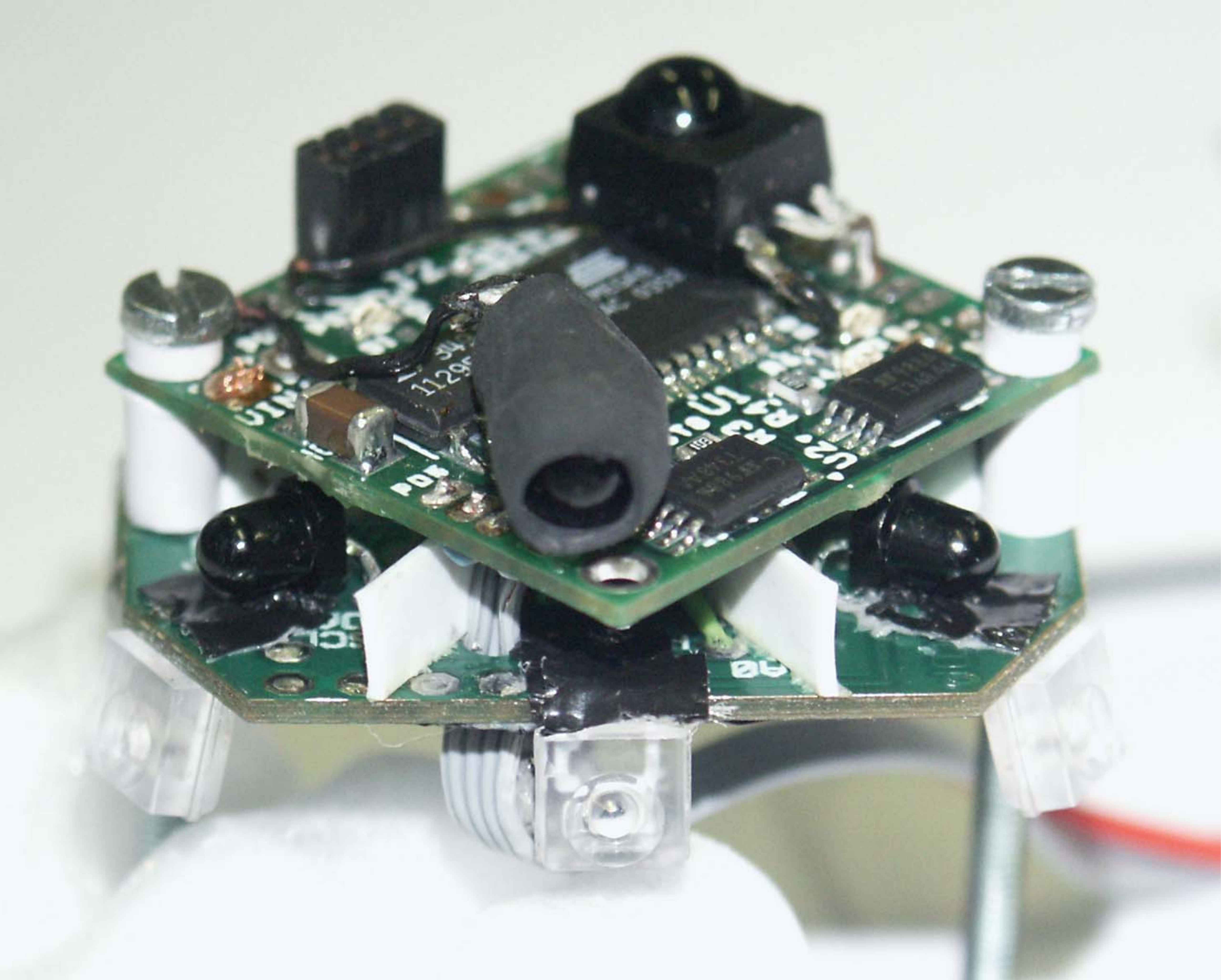}}\\
\subfigure[]{\includegraphics[width=0.49\textwidth]{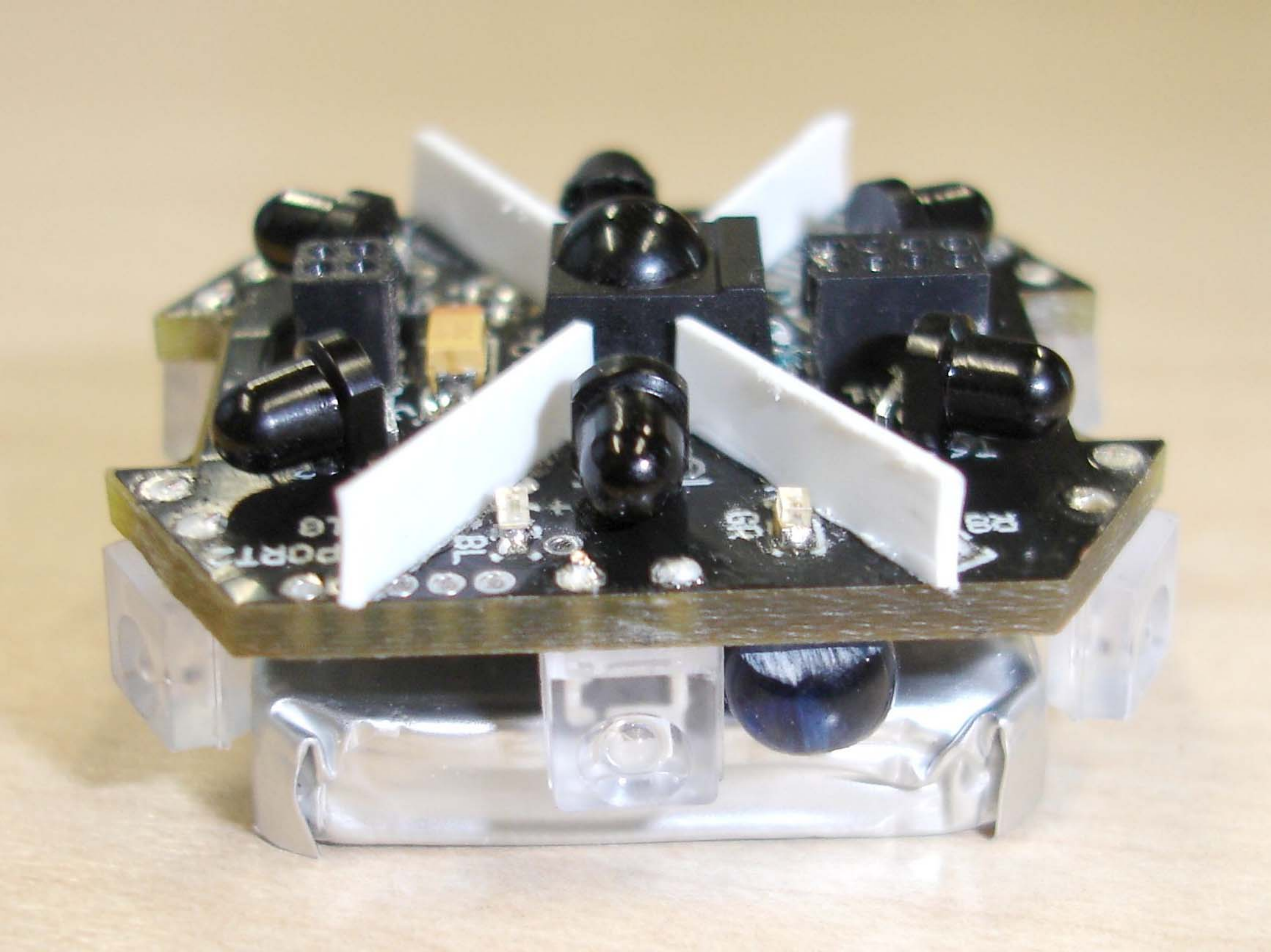}}
\caption {\small {\bf (a), (b), (c)} The first, second and third versions of the sensors board that support 6-x directional
robot-robot and host-robot communication proximity sensing and perception of surfaces geometry. \label{fig_expected_capab}
\label{fig:Jasmine} }
\end{figure}

Three versions of the sensors board are developed (Fig.~\ref{fig_expected_capab}), where we implemented different compromises between embodiment issues and requirements on communication radius $R_c$ and number of channels.

Generally, we tested over 30 pairs of IR receivers/emitters (as well as integrated devices) with 60$^\circ$ opening angle, corresponding sectoral coverage and communication distance. Analyzing the results of the tests~\cite{Kornienko_S05d}, we came to the conclusion that small integrated sensors are not really suitable for this application, although they have good coverage in 60$^\circ$ sector. The measured distances is only of 40-50 mm (on the brink of recognizability), and communication radius $R_c$ is about 60-70 mm (also on the brink of recognizability). The IR emitters with opening angle of $40^\circ$ and less do not provide a good coverage in 60$^\circ$ sector. From the tested IR emitters only one TSKS5400-FSZ demonstrated acceptable coverage that can be approximated in the algorithmic way. Together with with IR receiver TEFT4300, they build a receiver/emitter pair, used in all sensors boards. Receiver and emitter should be optically isolated so that to provide only 60$^\circ$ opening angle (they can perceive and send till 80-90$^\circ$). In the first prototype the optical isolation is done by putting small black tubes on receivers.

The communication signal from 150 mm distance on the direct line was of 0.7-0.8 V, in different directions within 60$^\circ$ not less than 0.1 V. The signal outside of 60$^\circ$ was less than 0.1 V for sensors with optical isolation. In this way robots can receive very exact information about a spatial origin of signal. Communication distance can be easy reduced (or even increased) by using submodulation or putting some threshold on the ADC values of sensors.

Tests of host-robot communication was performed by sending packages with PCM modulation. In remote control scheme, the input of PCM sensor (TSOP4836, 36kH subfrequency) is connected with the external interruption input of the microcontroller. Activating the interruption on the failing or rising edges we can differentiate between "T" and information impulses. Timer counts during information impulses so that we can easily recognize logical "0" and "1". Robot-robot communication utilizes similar principle, however does not modulate the signal with subfrequency. The duration of "T"-pulses was chosen to 1-0,5ms, so that at least the rate 1000 bit/sec can be provided.

The three developed versions of the sensors board differ in optical isolation, montage of sensors and electronics. Tubes on the receivers and montage of the first-version board restrict opening angle too much so that a large communication-dead zones appear in the corner areas, see Fig.~\ref{fig:problemProtocols}(a). In later sensors boards the receivers are placed on top side, emitters on the backside: in this way they are isolated by PCB. For providing 60$^\circ$ opening angle, all receivers are separated by plastic elements. Distance between receivers and a boundary of PCB determines minimal communication radius. After several optical simulation it was set to $\sim$2mm. Hardware recognition of communication signals demonstrated good results however was finally skipped due to size limitation.

There are implemented two communication protocols: with confirmation and without confirmation. For protocols with confirmation, each robot sends first a short request for communication (2x2ms impulses). When another robot gets this request, it sends a confirmation: "ready for communication" and waits for a package. Then the first robot sends 8 information and 1 parity bits. When this package is received, the robot confirms it by sensing: "package received". Both confirmations are implemented as 3x2ms impulses. For protocols without confirmation, a robot simply sends n-times the information package.

\subsection{Several Open Issues}

The imposed constraints on the microrobots are the main reason of appearing communication issues in a swarms: the smaller is the robot, the more limited is it~\cite{I-Swarm}. These constraints concern available energy (the most hardest issue), number, functionality and characteristics of on-board sensors and actuators, limited computational resources and specific micro-environment, where the robots operate in. The constrained swarm behavior possess very interesting properties:
almost all mechanisms of coordination, cooperation and communication work noticeably different to these mechanisms in
"usual large" systems. We can say, that they are so simplified, that even their physical embodiment into the robot plays enormous
role. In this section we discuss a few embodiment problems of communication in the micro-robot, e.g. RF via IR, ambient light,
recognition of signals and hardware-related protocols.

\textbf{1. RF via IR.} The required communication radius $R_{c}=50-140~mm$ can be implemented in the radio-frequency (RF) and infrared (IR) way. The \emph{RF} provides duplex communication within several meters and modern one-chip RF modules, even 802.11b/802.11g modules, consume energy in mW area. However we have a serious objection against RF in a swarm. Firstly, simultaneous transmissions of many (100+) micro-robots lead to massive RF-interferences. Secondly, RF-systems with a large communication radius transmit local information (exchange between neighbor robots) globally in a swarm. This local information does not have too much sense for all robots, so that we have a high communication overhead in this case. RF-communication is still useful for a global host-robot communication.

The \emph{IR communication} is recently dominant in so-called small-distance-domain, as e.g. for communication between laptops, hand-held devices, remote control and others. In IR domain we can choose between several different technologies, like IrDA. Additional advantage of IR solution consists in performing communication and proximity/distance sensing with the same sensors. IR emitter-receiver provides half-duplex communication, they are compact and energy consumption corresponds to I/O ports of microcontrollers. The IR solution is not new in robotic domain, see e.g.~\cite{Kube98}, however there are almost no solutions that combine perception, proximity sensing and communication.

The IR-equipment has also the problem of interferences. They appear, like in RF case, when several neighbor robots transmit
simultaneously. The problem of IR-interferences can be avoided by restricting an opening angle of a pair IR-receiver-transmitter.
For four communication channels, the opening angle of each channel is 90$^\circ$. In this case we have 2- and 3-robots
IR-interferences even in the "closest" radius (50 mm). Reducing the opening angle to 60$^\circ$ or to 40$^\circ$ allows avoiding
IR-interferences in the "close" and "near" radius (100 mm). Since many microcontrollers have 8-channel ADC (one ADC input is used by the distance sensor), we choose 6-channel directional communication.
\begin{figure}[ht]
\centering
\includegraphics[width=0.5\textwidth]{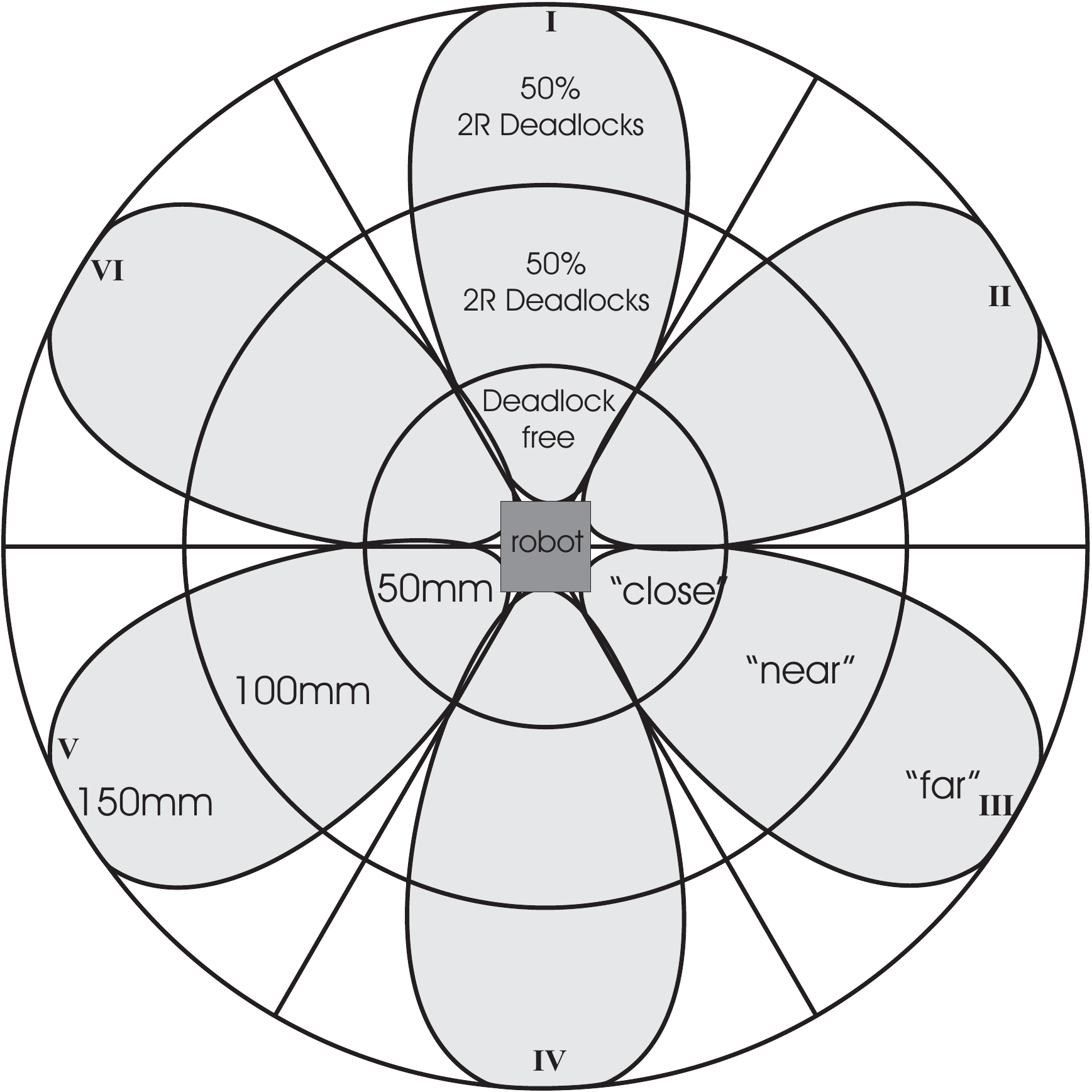}
\caption {\small Problem of IR-interferences in the "close", "near" and "far" communication zones. \label{fig:commRadius} }
\end{figure}

Directional communication is extremely important in a swarm also from another reason. The point is that a robot has to know not only a message itself, but also the context of this message (e.g. the direction from which the message is received, intensity of signal, communicating neighbor and so on). Without directional communication hardware, we cannot implement algorithms providing a spatial context~\cite{Kornienko_S05b}. From many software requirements the communication radius $R_c$ and the number of directional communication channels are the most important ones. From this viewpoint, the IR is more suitable for robot-robot communication than the RF.

\textbf{2. Influence of ambient light on communication/reflextion.} Speaking about IR communication we have to mention the problem of ambient light. Ambient light represents generally very critical issue, because it can essentially distort or even completely break IR communication/sensing. The experiments are performed with luminescent lamp, filament lamp and daylight. We can estimate three different components of a distortion introduced by ambient light. The direct light saturates photoelectric transistor so that it gets "blind". Secondly, ambient light reduces sensor sensitivity, even when it does not fail directly on sensor. Finally, indirect ambient light reduces contrasts between object and background, so that results of measurement are no more reliable and reproducible. So, a swarm has to be protected against a light of filament lamps. As far as possible, the direct daylight should be also avoided. Use of modulated light can essentially improve communication against ambient light, however this solution is not always feasible/acceptable.

The filament lamps can be used as a global signal to control a swarm~\cite{Bonabeau99}. When it is emitted simultaneously with
the luminescent light, the robot reacts more intensively on the filament light. This effect can be utilized in many purposes, like
finding the food source, navigation or even a quick message about some global event. This communication way does not require any additional sensors, however should be used only as an exception, because it essentially distorts a regular communication.

\textbf{3. Recognition of communication signals by hardware interruption.} The robot executes its own activities in a specific order: proximity sensing, receiving massages, sending messages, decision making and, finally, behavioral commands. One cycle of all these activities calls autonomy cycle. One autonomy cycle takes usually from 10ms till 150ms (it depends on communication and the number of elementary steps in each stage).
\begin{figure}[h]
\centering
\includegraphics[width=0.7\textwidth]{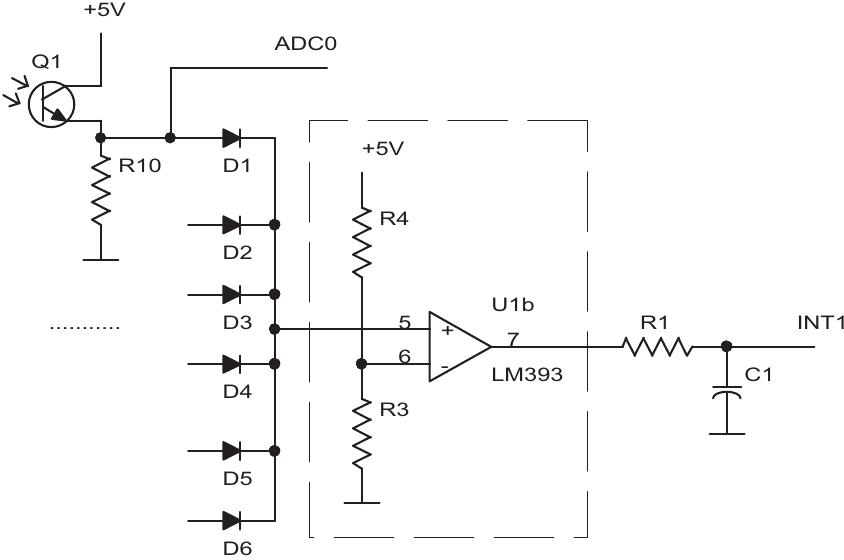}
\caption{\small Recognition of intensity and duration of communication pulses, as well as communication channel in the
hardware way. \label{fig: interuptions} }
\end{figure}

The robot scans one time per autonomy cycle all 6x-channels for the signal larger than a threshold. Performing 6x ADC conversions of 8 bit requires 1.2ms. When there is a signal (it could be a proximity or communication signal), the robot starts PCM encoding: it looks for "T" pulses by measuring the signal duration. The duration of proximity signals is 1ms, "T" pulses - 2ms. In this way the robot can easily filter out communication signals. In the worst case, robot has to wait 12ms+1.2ms to finish scanning all channels (without receiving any message). This is relatively large value, when to take into account that it is performed cyclically.

The ideas to recognize the intensity and duration of communication pulses in hardware way, and to call a hardware interruption, when such an impulse is received. The most simple way to do it consists in the differentiating RC filter (R1, C1 in Fig.~\ref{fig:
interuptions}). The required threshold can be set up by the divider R3, R4 in the comparator U1b. The corresponding channel
can also be encoded, when to put the signals before the diodes on the converter and then to digital port of microcontrollers.

\begin{figure}[ht]
\centering
\subfigure[]{\includegraphics[width=0.45\textwidth]{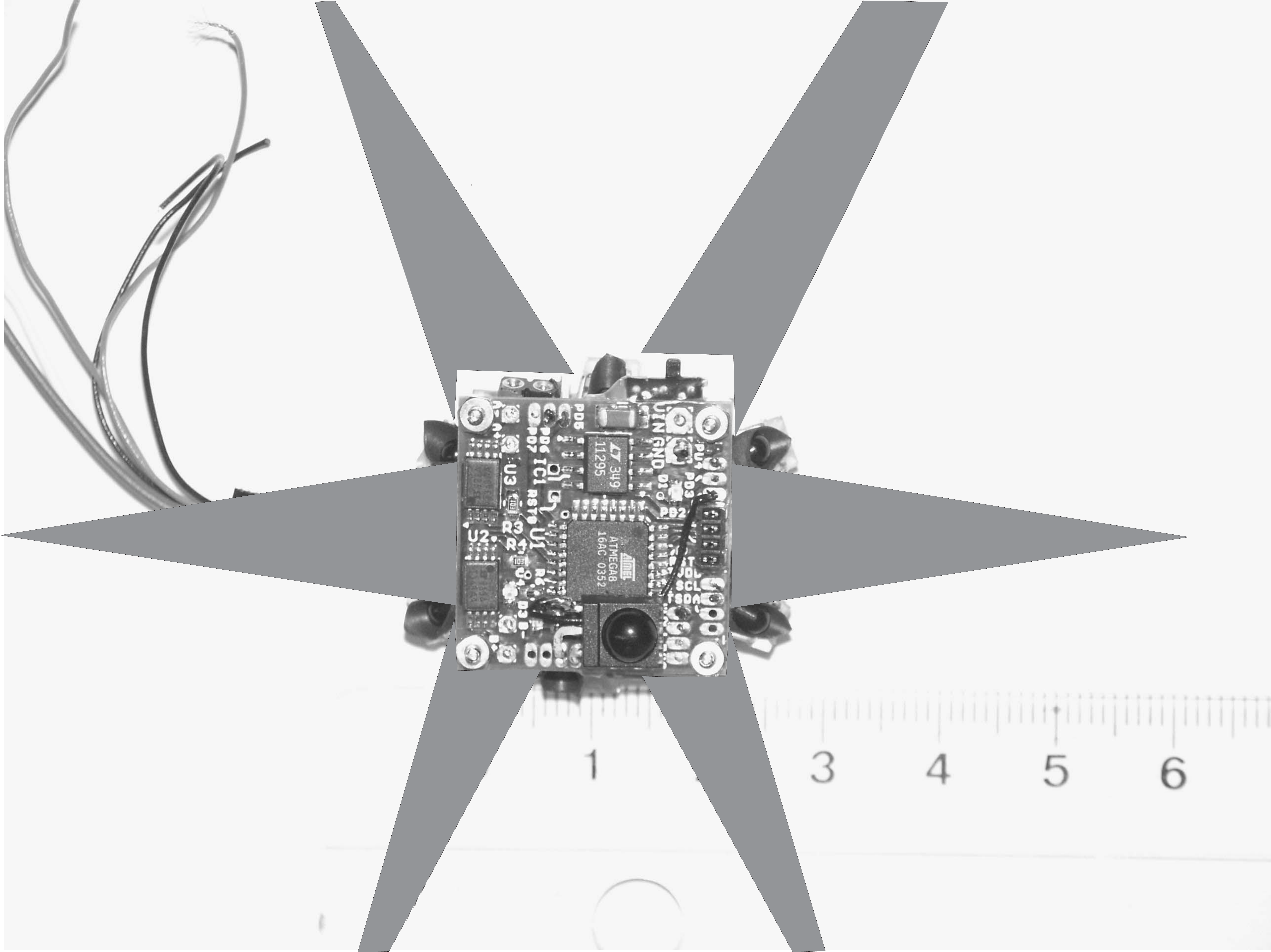}}~
\subfigure[]{\includegraphics[width=0.45\textwidth]{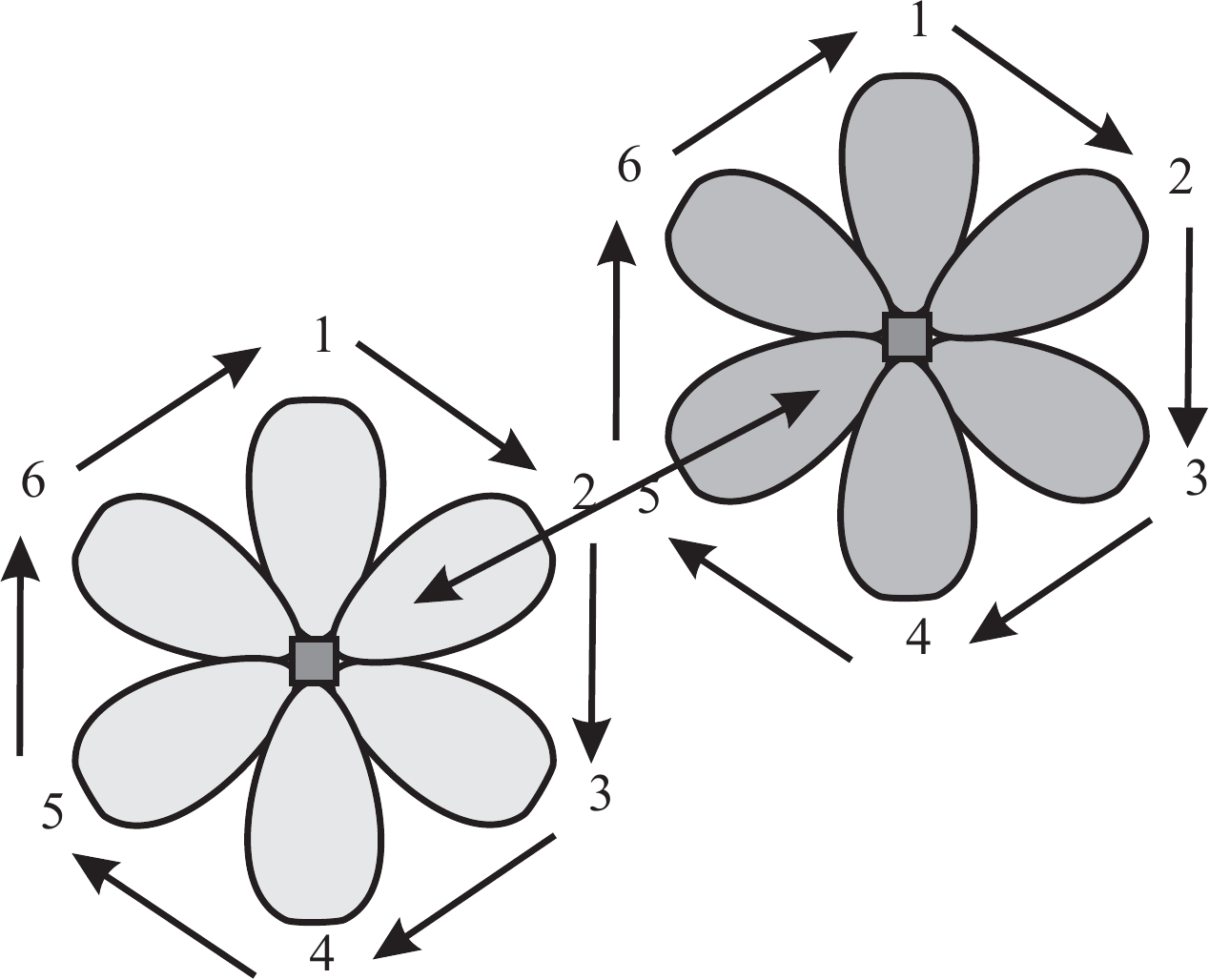}}\\
\subfigure[]{\includegraphics[width=0.7\textwidth]{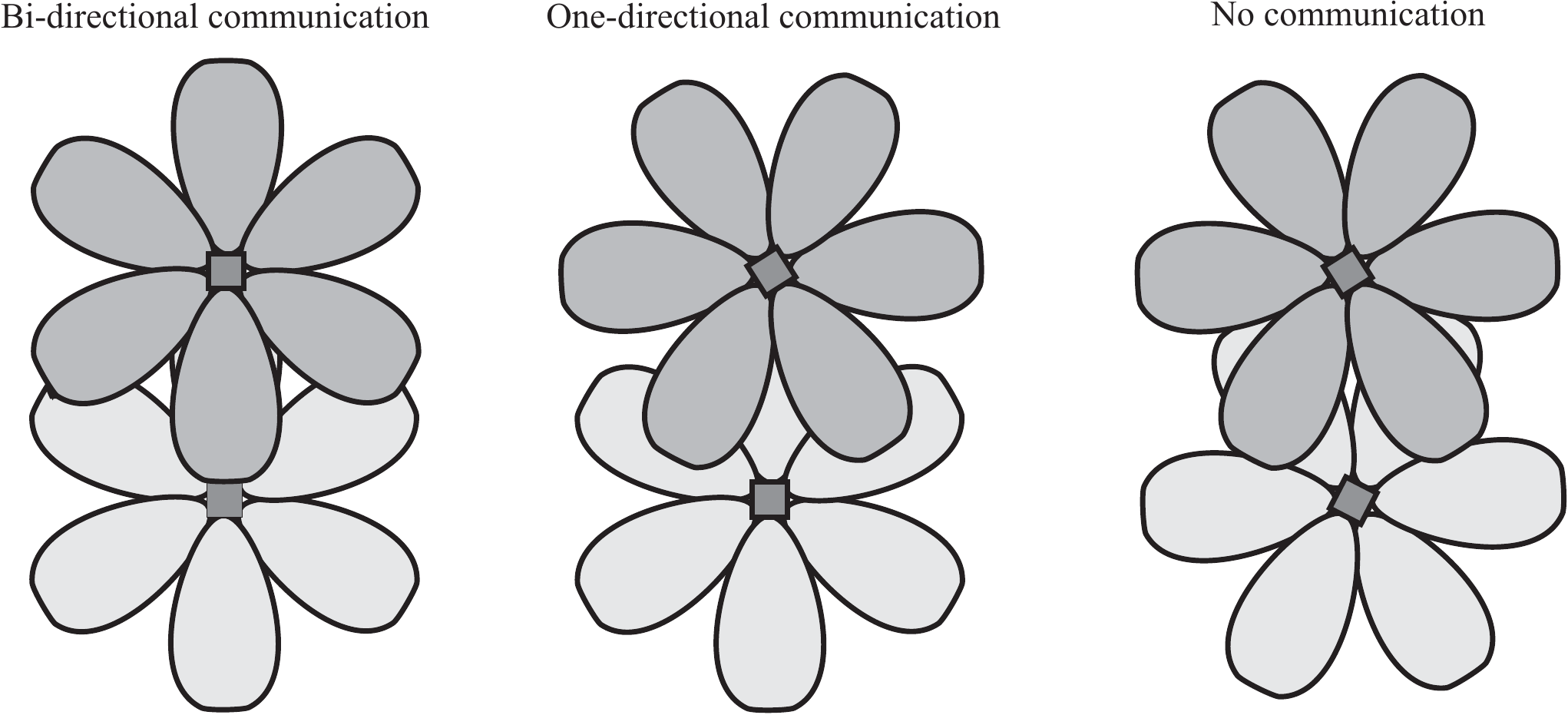}}
\caption{\small Problems of hardware-related protocols; {\bf (a)} Communication-dead-zones in the first version of the sensors board, shown as dark regions; {\bf (b)} Synchronization between receiving and sending channels, for bi-directional communication they should match; {\bf (c)} Nonlinear radiation patterns of receiver and sender. \label{fig:problemProtocols} \label{fig:commRadius1}}
\end{figure}

\textbf{4. Problems of hardware-related protocols.} The main problem we encountered in the first prototype of multi-channel
communication system is a poor probability of bi-directional communication contact, see Fig~\ref{fig:commRadius1}. 
The bi-directional communication is required for protocols with confirmation. The problem is not to have bi-directional communication: all robots can perfectly work in half-duplex mode (to send information in both directions). However in order to do this, robots have to be positioned into right positions and rotation angles. Bi-directional communication contact means that robots occupied corresponding positions and angles.

The reasons for poor bi-directional contacts are:\\[2mm]
- \textbf{appearance of communication-dead-zones} (primarily corners of the chassis, see Fig.~\ref{fig:problemProtocols}(a)) and the problem of emitter-receiver optical isolation that additionally increases these zones (they are different at emitters and receivers). We estimate that in average $\sim 10\%-15\%$ of the $360^\circ$
communication areal is lost;\\[2mm]
- \textbf{nonlinear radiation patterns} (Fig.~\ref{fig:problemProtocols}(b)). For bidirectional communication contact, both radiation patterns have to match. Comparing to one-directional communication, the probability of bidirectional contact on any arbitrary channel is 0.5-0.25 (according to the communication distance); \\[2mm]
- \textbf{the micro-robot can send and receive only sequentially} by all channels. In order to send a message, sending and receiving channels have to be "synchronized" (the number of a "sending" channel has to correspond to the number of a "listening" channel), see Fig.~\ref{fig:problemProtocols}(c). The probability that both channels "meet" is 1/6*1/6=1/36.\\[2mm]
Sending on one channel continues $\sim$ 38~ms for 8 bit package and is repeated each 10-100~ms (depend on the currently executed activities). With the probability of 1/36, the communication contact will be established within $p_t=\sim 1-1,5sec.$ and a
transmission of message (without confirmation) with 10 robots takes $ N p_t=\sim 10-15sec$. The transmission of messages with
the confirmation protocol takes $20-60sec.$ for the version I of the sensors board and $1-5sec.$ for the version II and III. These
data are confirmed by experiments~\cite{Pradier05},~\cite{Fu05}.

\section{Conclusion}
\label{sec:conclusion}

In this work we demonstrated a few hardware and software issues concerning IR-based perception and communication. We have clearly shown that several challenges created by limited hardware capabilities can be successfully resolved. The point of the demonstrated experiments is related to embodiment and to shifting information processing from a high-level symbolic and sub-symbolic representation to a low-level sensor-data processing. This should create a new inspiration towards reconfigurable and evolutionary approaches~\cite{Kernbach08Permis}, which utilize such low-level signals.

\small
%\bibliographystyle{plain}
%\bibliography{../../bibl_sk,../../own_bibl_sk}

\end{document}